\documentclass{article}

\usepackage{PRIMEArxiv}
\usepackage[T1]{fontenc}
\usepackage[ruled]{algorithm2e}
\usepackage{hyperref}
\usepackage{amsmath}
\usepackage{amssymb}
\usepackage{booktabs}
\usepackage{graphicx}
\usepackage{multirow}
\usepackage{float}
\usepackage{xcolor}
\usepackage{tikz}
\usepackage{xspace}
\usepackage{subcaption}
\usetikzlibrary{shapes.geometric}

\definecolor{stnstartcol}{HTML}{4DAF4A}  
\definecolor{stnmedcol}{HTML}{333333}    
\definecolor{stnendcol}{HTML}{377EB8}    
\definecolor{stnbestcol}{HTML}{E41A1C}

\newcommand{\stnmarkersize}{1.3ex}

\newcommand{\stnstart}{\tikz[baseline=-0.6ex]{%
  \node[draw=black, line width=0.4pt, fill=stnstartcol,
        rectangle, minimum size=\stnmarkersize, inner sep=0pt]{};}%
        \xspace}

\newcommand{\stntransition}{\tikz[baseline=-0.6ex]{%
  \node[draw=black, line width=0.4pt, fill=stnmedcol, fill opacity=0.5,
        circle, minimum size=\stnmarkersize, inner sep=0pt]{};}%
        \xspace}

\newcommand{\stnend}{\tikz[baseline=-0.6ex]{%
  \node[draw=black, line width=0.4pt, fill=stnendcol,
        regular polygon, regular polygon sides=3,
        minimum size=\stnmarkersize+0.2ex, inner sep=0pt]{};}%
        \xspace}

\newcommand{\stnbest}{\tikz[baseline=-0.6ex]{%
  \node[draw=black, line width=0.4pt, fill=stnbestcol,
        circle, minimum size=\stnmarkersize, inner sep=0pt]{};}%
        \xspace}
\begin{document}
%
\title{Semantic Space Search Trajectory Networks}


%
%

\author{
Julian Agudelo\\
UMR MIA-PS and Agrial\\
AgroParisTech \\
Palaiseau, France \\
\And
Alberto Tonda\\
UMR MIA-PS and ISC-PIF \\
INRAE and CNRS \\
Palaiseau, France \\
\And
Gabriela Ochoa \\
Computing Science and Mathematics \\
University of Stirling \\
Scotland, United Kingdom \\
\And
Vincent Guigue \\
UMR MIA-PS \\
AgroParisTech \\
Palaiseau, France \\
\And
Cristina Manfredotti \\
UMR MIA-PS \\
AgroParisTech \\
Palaiseau, France \\
\And
Evelyne Lutton \\
UMR MIA-PS and ISC-PIF \\
INRAE and CNRS \\
Palaiseau, France \\
}


\maketitle 
\begin{abstract}
Search Trajectory Networks (STNs) are a graph-based tool for visualizing and characterizing the behavior of optimization algorithms. STNs' reliance on discretization of the search space has largely confined them to low-dimensional or combinatorial settings. We introduce a methodology for constructing STNs in semantic spaces, defined as the space of a model's predictions on a fixed sample set. Our approach discretizes semantic vectors and aggregates them into network nodes via agglomerative clustering with complete linkage under a normalized Hamming distance. Since any predictor can be summarized by its semantic vector, this method enables comparison of learning dynamics across otherwise incomparable algorithm families. We apply semantic space STNs to classification and regression tasks solved using different machine learning algorithms, recovering known qualitative differences between them. Additionally, we use semantic space STNs to study neural network generalization by contrasting standard training with the label randomization regime of Zhang et al.~\cite{zhang2017rethinking}. The resulting STNs exhibit consistent structural differences, training on real labels produces denser, more efficient and more centralized graphs than training on shuffled labels. Together, our results show that semantic space STNs capture functional training dynamics arising from the interaction between learning algorithms and data, providing a tool for analyzing and comparing learning dynamics across machine learning models and training regimes.

\keywords{Search Trajectory Networks  \and Semantic space \and Machine Learning \and Deep Neural Networks Generalization}
\end{abstract}
%
%



\section{Introduction}

Training a machine learning model is fundamentally equivalent to using an optimization algorithm to traverse a space of candidate models in pursuit of one which fits the data the best.
The path taken to reach the final solution—how the search explores, where it concentrates, and which regions act as attractors—carries information that metrics cannot convey. Characterizing this behavior is the object of fitness landscape analysis, which studies the structure of the surface an optimizer navigates and the dynamics it induces.

Search Trajectory Networks (STNs)~\cite{ochoa2021search} are a graph-based tool within this tradition. They represent optimization as a network of representative states connected by observed transitions and aggregated over multiple runs. However, their construction relies on discretization of the search space, which becomes problematic in high-dimensional settings due to distance concentration and the difficulty of defining meaningful neighborhoods~\cite{beyer1999when,goos2001surprisingdistance}.

A complementary perspective is provided by semantic space representations~\cite{moraglio2012geometric}, which describe models by their behavior rather than their parameters, using prediction vectors over a fixed dataset. This yields a common representation for heterogeneous learners and induces a functional notion of similarity between models.

We combine these ideas to construct Semantic Space STNs. This removes the dependence on explicit search space discretization and enables comparison of learning dynamics across otherwise incomparable algorithm classes. Our contributions are:
\begin{enumerate}

    \item We introduce a clustering-based discretization strategy that enables the construction of STNs in high-dimensional semantic spaces.

    \item We use Semantic Space STNs to perform qualitative and quantitative comparisons of learning dynamics across algorithm families.
    
    \item We use Semantic Space STNs to characterize standard training and the label randomization regime of Zhang et al.~\cite{zhang2017rethinking}, revealing systematic differences in learning dynamics and offering an alternative perspective on the study of generalization in neural networks.
\end{enumerate}




    

\textbf{The code is publicly available at:} \href{https://github.com/JulAgu/SSSTN}{https://github.com/JulAgu/SSSTN}

\section{Background}

\textbf{Search Trajectory Networks (STNs)} are a graph-based representation model for visualizing and characterizing the search behavior of an optimization process, in which nodes correspond to states of the search process and edges encode transitions between consecutive states \cite{ochoa2021search}. Formally, a STN is a graph $G=(N,E)$ with $N$ the set of nodes and $E$ the set of edges. \textit{Nodes} are non-empty subsets of solutions obtained from a predefined discretization of the search space. \textit{Edges} are directed and link consecutive states along the search trajectory. They are weighted by the frequency with which a transition between two nodes occurs during optimization.

An STN is constructed from trajectories generated during multiple independent executions of an algorithm on a given problem instance. We record each algorithm iteration as a sequence of steps connecting two adjacent solutions in the search process. The representative solutions constitute the nodes of the network, whereas transitions between consecutive solutions define its directed edges. For each solution, both its localization (solution vector) and fitness value are stored.

Using a collection of trajectories, a post-processing step aggregates information from all runs to construct a single network. Solutions corresponding to the same representative location in the search space are mapped to a unique node according to a problem-dependent representation. The resulting STN allows for both qualitative and quantitative analyses, providing insights into the behavior of an algorithm. Several extensions of STNs have been proposed and used for a wide range of purposes, including analyzing the behavior of metaheuristics~\cite{ochoa2021search}, analyzing multiobjective evolutionary algorithms~\cite{jimenez2022multiobjective}, analyzing multiobjective combinatorial optimization~\cite{perez2023combinatorial}, and visualizing combinatorial spaces of neuroevolutionary algorithms~\cite{rodrigues2022stnneuralnets}.



The term \textbf{Semantic Space} is independently used by different research communities, with different meanings. For example, in deep learning, it refers to embeddings in the latent space; in evolutionary computation, it represents the space defined by vectors containing the predictions of different machine learning models (typically GP trees) over a fixed set of samples. Here, we adopt the evolutionary computation interpretation of the term, formally defined below.

Given an ordered set of sample covariates $\mathbf{x} = \{x^{(0)},\dots,x^{(N)}\}$ and a predictor $f$, the \emph{semantic vector} $\mathbf{\hat{y}} = f(\mathbf{x}) = \{\hat{y}^{(0)},\dots,\hat{y}^{(N)}\}$ contains all of $f$'s predictions for the samples in $\mathbf{x}$. The \emph{target semantic vector} $\mathbf{y}$ contains the ground-truth values for the samples, with $\mathbf{\hat{y}} = \mathbf{y}$ if $f$ makes perfect predictions. The relative positions of two predictors in the semantic space carry more information than their aggregate metrics~\cite{fenoglio2024federated}. For an iterative learner, the sequence $\langle \mathbf{\hat{y}}_{0}, \dots, \mathbf{\hat{y}}_{T} \rangle$ over $T+1$ optimization steps traces a trajectory whose geometry encodes the behavior of learning. Most modern machine learning algorithms are iterative learners, ranging from neural networks (learning over epochs) to boosted trees and symbolic regression. We argue that semantic spaces are particularly meaningful when studying the behavior of these algorithms.


Semantic spaces were first introduced in Geometric Semantic Genetic Programming~\cite{moraglio2012geometric}, an evolutionary technique that aims to perform mutations always moving individuals toward the target semantic vector $\mathbf{y}$. Semantic spaces are typically high-dimensional, with $|\mathbf{y}| \approx [10^2, 10^5]$. When dimensionality reduction is applied, either using techniques such as MAP-Elites~\cite{mouret2015illuminating,zhang2023mapelites} or to allow humans to visually inspect results~\cite{fenoglio2024federated}, the resulting reduced space is sometimes called \emph{behavioral space}.
%
%
%
\section{Building STNs in High-Dimensional Semantic Spaces}
\label{sec:methodology}

We consider supervised machine learning tasks, specifically classification and regression, in which the objective is to learn a mapping from inputs to target outputs using different algorithms.
%
Our objective is to build Search Trajectory Networks (STNs) in the semantic space. This choice enables direct comparison of training dynamics across heterogeneous algorithms that are otherwise not comparable. For example, the weights of a neural network, the trees of a gradient-boosted ensemble, and the expressions evolved by symbolic regression cannot be meaningfully aligned. However, at any point during optimization, all of these models can be represented by a common object: a semantic vector containing their predictions on the training set.

Given a training dataset $\mathcal{D} = \{(x^{(i)}, y^{(i)})\}_{i=0}^{N}$, a machine learning algorithm yields a sequence of semantic vectors during optimization. At optimization step $t$, the algorithm induces a semantic vector $\mathbf{\hat{y}}_t = \big(f_t(x_0), \ldots, f_t(x_N)\big)$, where $f_t$ denotes the predictor obtained at that step. In this way, the sequence $\langle \mathbf{\hat{y}}_{0}, \dots, \mathbf{\hat{y}}_{T} \rangle$ defines the search trajectory in the semantic space. As we are interested in the behavior of the optimization process rather than in a particular realization, we perform multiple training runs under different initial conditions, as it is typical in the construction of STNs~\cite{ochoa2021search}.

Building an STN requires mapping semantic vectors to nodes via a location function. This function determines when two vectors should be considered as representing the same functional state of the learning process, effectively defining an aggregation of the underlying space. A central difficulty stems from the fact that this aggregation is not naturally well defined in high dimensions. Even when operating in a semantic space that is of a lower-dimension than some parameter space (such as that of a neural network), its effective size remains large, and local neighborhoods tend to be sparse. As a consequence, classical notions of proximity become unstable due to several phenomena related to the curse of dimensionality: distances concentrate~\cite{goos2001surprisingdistance}, small perturbations can induce disproportionately large changes in nearest-neighbor structure~\cite{beyer1999when}, and clustering outcomes become highly sensitive to the choice of metric.
%
%
%
%
To address these limitations, we propose an approach based on agglomerative clustering with complete linkage and normalized Hamming distance. Specifically, we first collect the discretized semantic vectors observed across optimization runs. Each vector is initially treated as its own cluster. We then iteratively merge the pair of clusters with the smallest complete-linkage (farthest-neighbor) Hamming distance, defined as the maximum pairwise Hamming distance between elements of the two clusters. Merging continues until the smallest inter-cluster distance exceeds a threshold $\tau$, or only one cluster remains. The resulting clusters define the nodes of the STN, while transitions between consecutive semantic vectors are mapped to edges between their corresponding cluster assignments. The construction of semantic space STNs for classification and regression tasks differs only in how they are discretized to enable the computation of the Hamming distance.




\textit{Classification tasks} Our method requires each semantic vector to be a vector of predicted class labels. While some learning algorithms naturally produce a single label prediction for each instance, others, such as neural networks, output a vector of logits\footnote{In deep learning terminology, \textit{logits} are the unnormalized scores produced by the final layer of a neural network before being transformed into class probabilities in a classification problem.} whose dimension equals the number of classes $C$. In these cases, the prediction matrix $\mathbf{\hat{y}} \in \mathbb{R}^{N \times C}$ is converted into a label vector $\mathbf{\hat{y}} \in \{1,\ldots,C\}^N$ by applying the $\mathrm{argmax}$ operator along the class dimension. It is precisely this resulting vector that we call the semantic vector in the case of classification.

\textit{Regression tasks} For regression, predictions are continuous and thus not directly comparable using Hamming distance. We therefore discretize the semantic vectors by partitioning prediction values into global quantile-based bins computed over all semantic vectors. Each prediction vector is then mapped to a corresponding vector of bin assignments, yielding a discrete representation analogous to the classification case.

Once we have the set of trajectories of semantic vectors $\{\langle \mathbf{\hat{y}}_{0}, \dots, \mathbf{\hat{y}}_{T} \rangle^{(r)}\}_{r=1}^{r=R}$ for $R$ runs, we apply the agglomerative clustering. Finally, we broadcast the cluster identifiers, which define the nodes, back to each semantic vector, and count the number of transitions between nodes to define the edges, thereby obtaining the STN. Conveniently, the only hyperparameter in this methodology is the threshold $\tau$ for agglomerative clustering. This parameter controls the level of aggregation in the resulting dendrogram and can be adjusted manually to explore different granularities of the STN. At the same time, $\tau$ admits data-driven selection strategies, suggesting possible automation of selection based on stability or structure-based criteria, which we leave for future work.

\textbf{On Methodological Design Choices} Our method operates in the space of discretized model predictions rather than in the underlying continuous logit or regression output space, although both can be regarded as variations of the semantic space. This design choice is motivated by the objective of STNs, which is to capture stable states of the optimization process. We observe that the logit or regression output space forms a continuous, uncalibrated representation in which distances are sensitive to scaling and inter-class coupling, making it unsuitable for defining such states. Importantly, using the space of discretized model predictions does not restrict the method to models that produce logits, as many machine learning algorithms in classification settings directly output discrete labels without an intermediate continuous representation.

After discretization, each semantic vector lies in a finite categorical space where similarity is naturally expressed as coordinate wise agreement. In this setting, Hamming distance provides a direct measure of disagreement between states and is invariant to any implicit ordering of class labels, unlike Euclidean or cosine distances applied to logits.

We employ agglomerative clustering because it naturally builds a hierarchy of merges based on pairwise similarity, allowing stable discovery of recurring semantic configurations across runs and providing a method depending on only one hyperparameter, the threshold $\tau$. This combination yields a structure-preserving partition of the space, which is essential for constructing STNs.





\section{Analyzing Supervised ML Algorithms}
\label{sec:algo_comparaison}

To validate the proposed STN construction methodology, we conduct experiments on benchmark classification and regression tasks using three representative learning algorithms: multi-layer perceptrons (MLPs), XGBoost, and symbolic regression (SR). MLPs and XGBoost are applied to both regression and classification problems, while SR is limited to regression tasks, since this technique cannot directly handle classification problems with more than two classes\footnote{Although classification problems can be cast as regression by training one SR model per class and applying a softmax to decide which prediction wins, this extension is not considered in this work.}.
%
%
Table~\ref{tab:datasets} introduces the datasets used in this study. We note that the number of training examples equals the dimensionality of the semantic space. Table~\ref{tab:used_models} presents the learning algorithms used in this study.

\begin{table}[th]
\centering
\footnotesize
\begin{minipage}[b]{0.53\textwidth}
  \centering
  \caption{The datasets used. The top and bottom block correspond to classification and regression task
  . The first column lists the corresponding OpenML IDs.}
  \label{tab:datasets}
  \resizebox{!}{5.7em}
  {
  \setlength{\tabcolsep}{2pt}
  \begin{tabular}{@{}r l c c r@{}}
    \toprule
    did & Dataset & Cls & Feat. & \#Train \\
    \midrule
    4134  & Bioresponse       & 2  & 1776 & 3375  \\
    40927 & CIFAR 10          & 10 & 3072 & 54000 \\
    40996 & Fashion-MNIST     & 10 & 784  & 63000 \\
    554   & MNIST (784)       & 10 & 784  & 63000 \\
    \midrule
    44994 & cars              & -- & 17   & 723   \\
    44978 & cpu activity      & -- & 21   & 7372  \\
    44960 & energy efficiency & -- & 8    & 691   \\
    \bottomrule
  \end{tabular}
  }
\end{minipage}
\hfill
\begin{minipage}[b]{0.43\textwidth}
  \centering
  \caption{Algorithms and their configurations. MLPs and XGBoost are applied across all datasets, while SR is used  for regression.}
  \label{tab:used_models}
  \resizebox{!}{5.7em}
  {
    \setlength{\tabcolsep}{6pt}  
    \begin{tabular}{l l}
        \toprule
        Algorithm & Configuration \\
        \midrule
        \multirow{2}{*}{MLP$_{\mathrm{cls}}$} & $3{\times}512$, SGD \\
                                              & 1000 epc, lr = 5e-2 \\
        \midrule
        \multirow{2}{*}{MLP$_{\mathrm{reg}}$} & $2{\times}512$, SGD \\
                                              & 5000 epc, lr = 1e-3 \\
        \midrule
        \multirow{2}{*}{XGBoost} & 1000 rounds, depth 6 \\
                                 & $\eta{=}0.1$, subsample 0.8 \\
        \midrule
        SR & 40 gen., $\{+,-,\times,\div\}$ \\
        \bottomrule
    \end{tabular}
  }
\end{minipage}
\end{table}

For each algorithm-dataset pair, we generate a collection of runs that differ in their stochastic initialization to induce diversity in the explored semantic space. The source of stochasticity depends on the learning algorithm. For MLPs, it corresponds to the random initialization of network weights. For XGBoost, it is the random seed controlling row and column subsampling. For SR, it is the random seed governing the initialization of the population of candidate expressions and the stochastic evolutionary operators used during the search process. The dataset, train/test splits, and hyperparameters are kept fixed across runs. Since our goal is not to optimize performance, we do not perform dataset-specific hyperparameter tuning. Table~\ref{tab:results} presents the results of the experimental campaign.

\begin{table}[th]
    \centering
    \caption{Test set fitness across datasets. Regression uses $R^2$ and classification uses accuracy. We report mean and standard deviation over 30 runs.}
    \label{tab:results}
    \setlength{\tabcolsep}{4pt}
    \begin{tabular}{l l c c c}
        \toprule
        Dataset & Metric & MLP & XGBoost & Symbolic Regression \\
        \midrule

        Bioresponse     & \multirow{4}{*}{Accuracy}
                          & $0.7933 \pm 0.0091$ & $\mathbf{0.8067 \pm 0.0048}$ & - \\
        CIFAR\_10       & & $0.4896 \pm 0.0048$ & $\mathbf{0.5937 \pm 0.0039}$ & - \\
        Fashion--MNIST  & & $0.8878 \pm 0.0023$ & $\mathbf{0.9117 \pm 0.0009}$ & - \\
        mnist\_784      & & $0.9759 \pm 0.0009$ & $\mathbf{0.9818 \pm 0.0005}$ & - \\
        \midrule

        cars                & \multirow{3}{*}{$\mathrm{R2}$}
                              & $\mathbf{0.9608 \pm 0.0025}$ & $0.9521 \pm 0.0025$ & $0.8958 \pm 0.0148$ \\
        cpu activity       & & $0.9781 \pm 0.0012$ & $\mathbf{0.9878 \pm 0.0003}$ & $0.9636 \pm 0.0073$ \\
        energy efficiency  & & $0.9964 \pm 0.0003$ & $\mathbf{0.9978 \pm 0.0001}$ & $0.9087 \pm 0.0201$ \\
        \bottomrule
    \end{tabular}
\end{table}

\begin{figure}[htbp]
    \centering
    \includegraphics[width=0.45\textwidth]{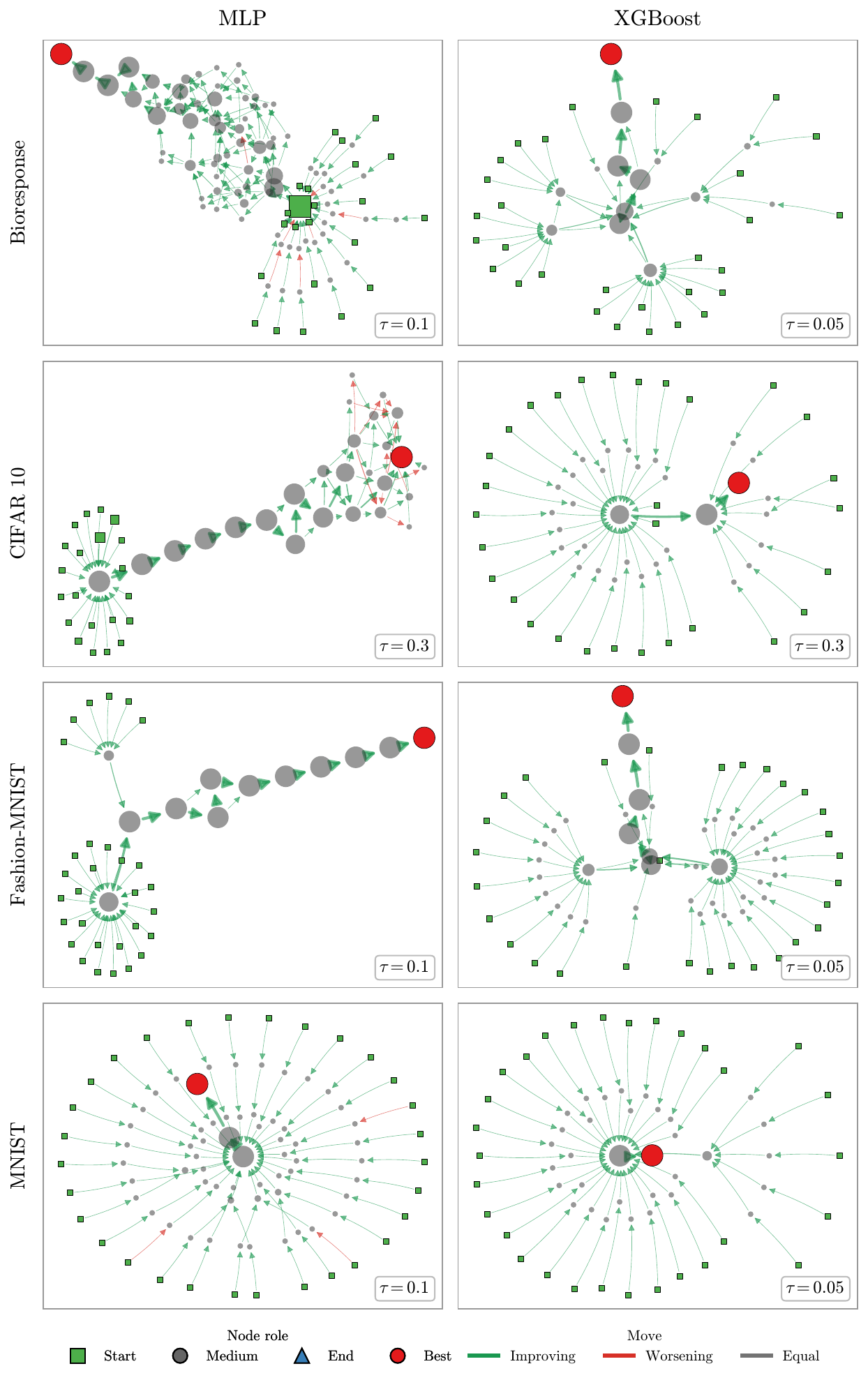}
    \caption{Semantic space STNs for MLP and XGBoost on the four classification datasets, aggregated over 30 runs. The clustering threshold $\tau$ was selected manually per dataset for readability.}
    \label{fig:comparison_classification}
    \vspace{3mm}
    \centering
    \includegraphics[width=0.63\textwidth]{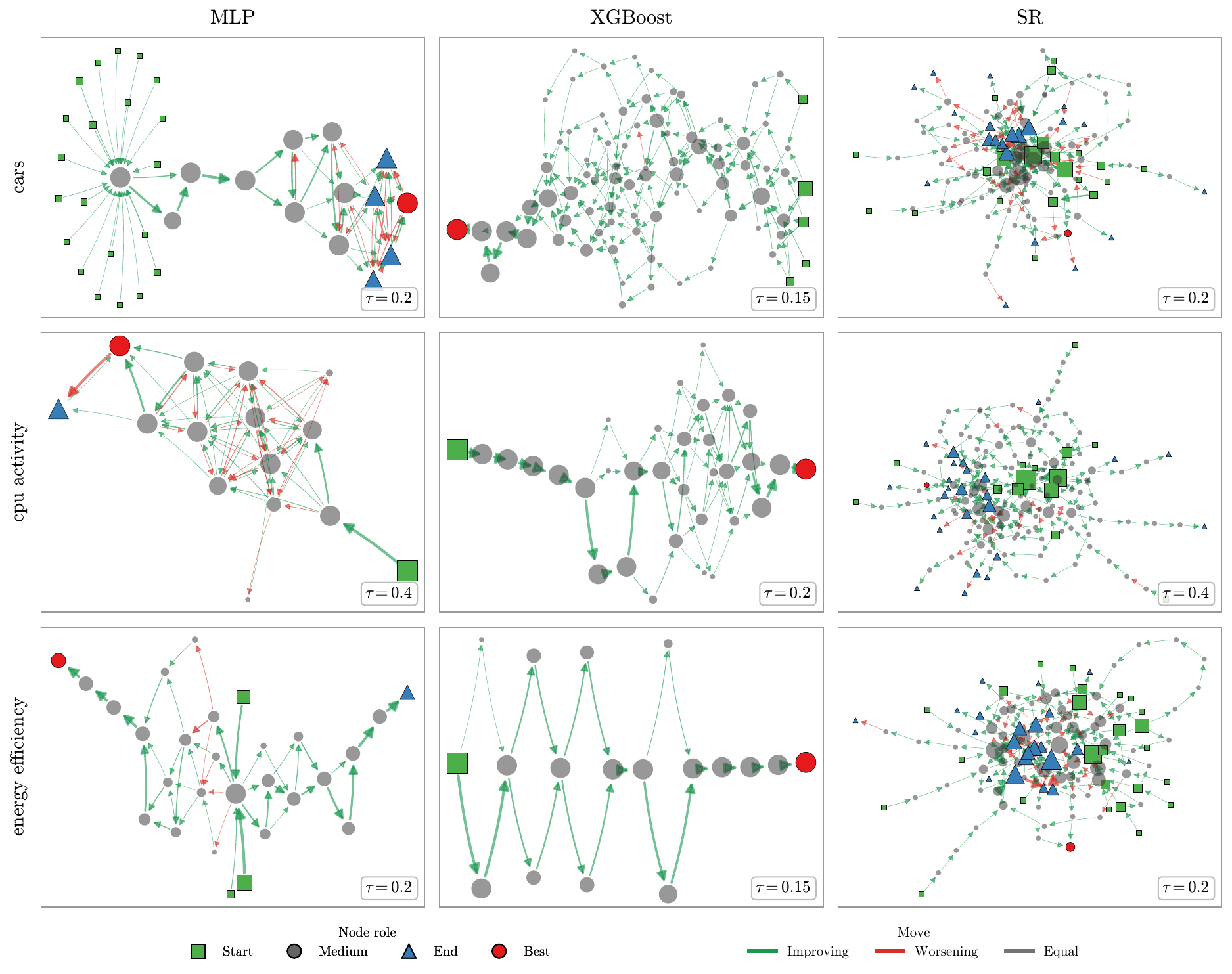}
    \caption{Semantic space STNs for MLP, XGBoost and Symbolic Regression on the three regression datasets, aggregated over 30 runs. Continuous predictions are discretised into 10 global quantile bins before Hamming-based clustering.}
    \label{fig:comparison_regression}
\end{figure}

Figures \ref{fig:comparison_classification} and \ref{fig:comparison_regression} show the STNs corresponding to the aforementioned experiments. The node size encodes the number of semantic vectors representing the state, while the edge width reflects the number of transitions. Unlike the original STN formulation, which defines the end node as the last newly visited state, we use the true terminal state because training trajectories often revisit earlier nodes before terminating.
We adopt the visual conventions of prior STN works, including start (\stnstart), transition (\stntransition), end (\stnend), and best (\stnbest) nodes, which contain the semantic vector with the highest fitness. Furthermore, we introduce directed edges to indicate the direction of optimization and color them to show whether a transition corresponds to an improvement or a deterioration in the fitness value. The figures are generated using the Kamada-Kawai layout algorithm~\cite{kamada1989algorithm}.

Figure~\ref{fig:comparison_classification} highlights the striking similarities between the STNs of MLP and XGBoost on classification tasks. Since both algorithms optimize a loss function by following its gradient, multiple starting nodes (corresponding to random weight initialization for MLP and a first decision tree built on a random subsample of the training data for XGBoost) quickly converge toward a common area of the semantic space (representing a funnel-like transition region), then end up in a final region of the semantic space, visualized as the best node \stnbest, and likely corresponding to slight overfitting of the training set, close to perfect prediction. Interestingly, the STN for MLP on CIFAR 10 is the only case where the graph presents a set of transition nodes \stntransition strongly connected to the best node \stnbest. Looking back at Table~\ref{tab:results}, CIFAR 10 also shows the worst performance of MLP, possibly indicating that the model capacity used in the experiments is insufficient to achieve good accuracy, thus likely affecting the convergence of the stochastic gradient descent algorithm. For other cases, where both algorithms perform well, the main difference in the trajectories seems to be the number of common transition nodes (\stntransition), so that for MLP any search is quickly directed toward this common region and then proceeds along the same path, while for XGBoost only the last few steps are common across trajectories.

Figure~\ref{fig:comparison_regression} depicts a similar pattern. On regression tasks, MLP and XGBoost converge toward a common region of the semantic space, after which MLP converges to what appears to be a basin of attraction, where multiple transition nodes \stntransition and the best node \stnbest are strongly interconnected. On the other hand, XGBoost consistently converges to a single best node \stnbest, exhibiting a largely deterministic optimization process that begins from a shared constant prediction and greedily fits residual gradients. Limited randomness introduced through subsampling acts only as a perturbation around a single attractor, causing runs to converge to nearly identical prediction functions.

Notably, MLP basins of attraction (a group of nodes that are closely interconnected and located near a best node \stnbest) include worsening transitions that could indicate a set of mutually reachable, near-degenerate solutions among which trajectories transit. The energy efficiency task is the exception: two terminal nodes (\stnend and \stnbest) lie at the tips of separate trajectory branches. We hypothesize that this reflects two genuinely distinct basins.
%

SR exhibits a distinct behavior, with STNs forming a markedly
different structure from the previous two algorithms and reflecting more exploratory dynamics in semantic space, with end nodes, \stnend, distributed across a wide range of locations. In contrast to MLP and XGBoost, SR explores semantic space in multiple directions rather than converging along a common trajectory. This behavior is consistent with conventional wisdom regarding how these two types of algorithms work: gradient-based methods move in a defined direction dictated by the loss gradient, while symbolic regression explores the space more freely through stochastic genetic operators rather than following a single optimization direction.

\begin{figure}[htbp]
    \centering
    \includegraphics[width=\textwidth]{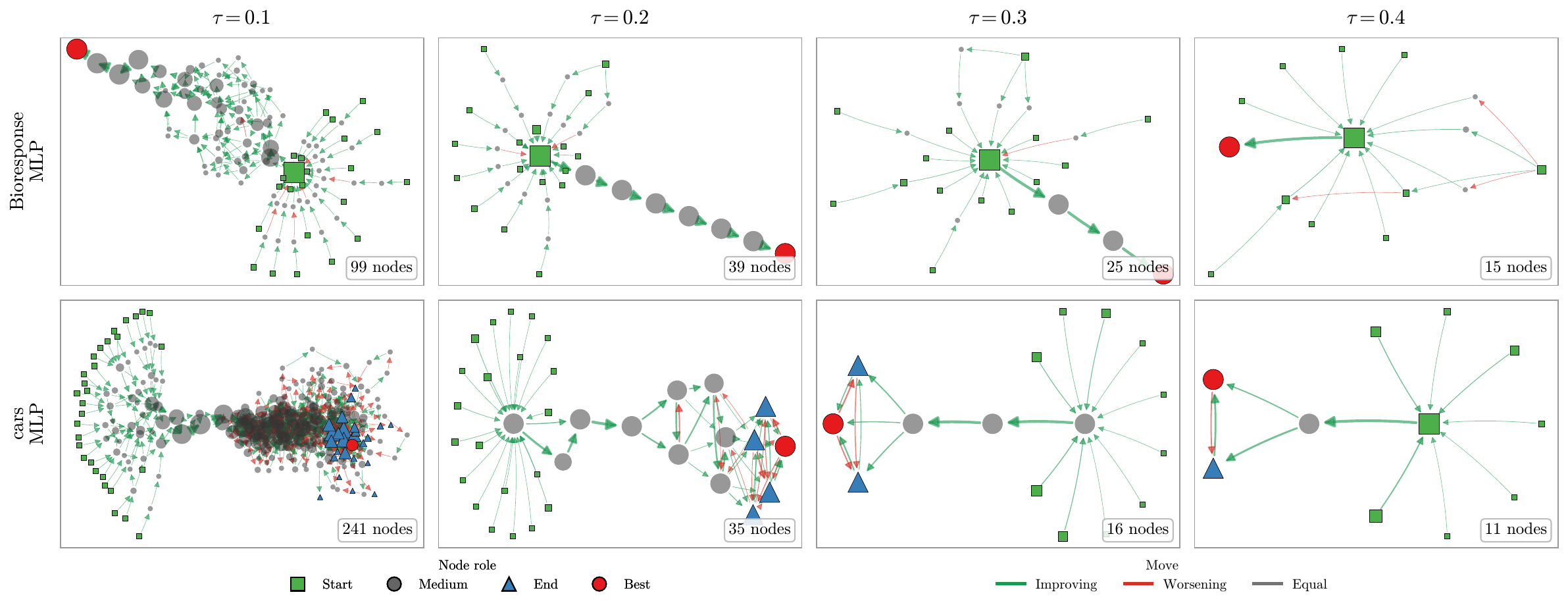}
    \caption{Effect of the agglomerative clustering threshold $\tau$ on STN granularity. Increasing $\tau$ coarsens the aggregation, merging semantic states and exposing the optimisation dynamics at progressively higher levels of abstraction.}
    \label{fig:threshold_sweep}
\end{figure}

As described in Section~\ref{sec:methodology}, the only hyperparameter of our method is the agglomerative clustering threshold $\tau$. Figure~\ref{fig:threshold_sweep} shows how varying this threshold affects the granularity of the resulting aggregation. By controlling the level of abstraction of the STN, $\tau$ provides a practical mechanism to explore the hierarchy of optimization dynamics.

\begin{figure}[htbp]
    \centering
    \includegraphics[width=0.8\textwidth]{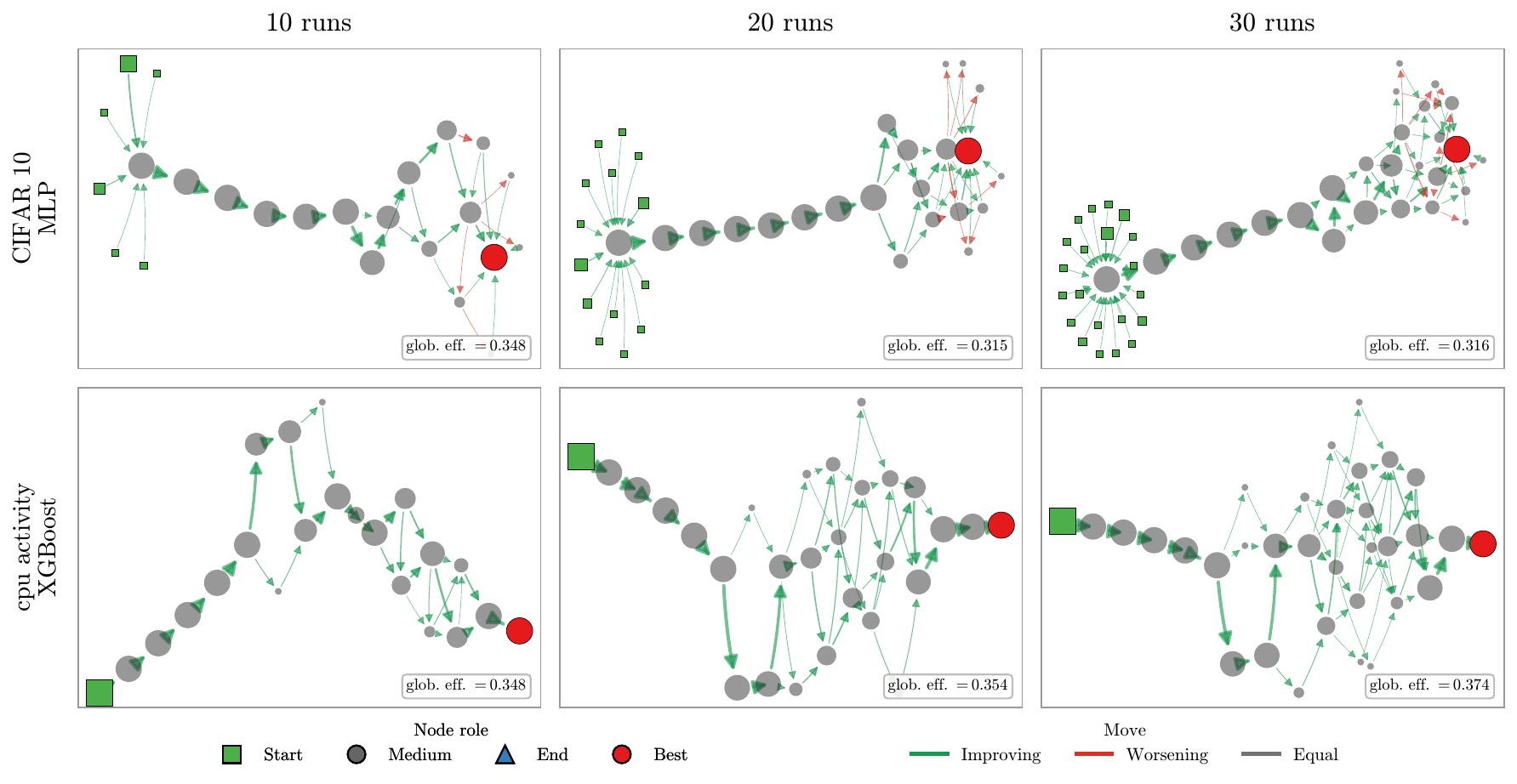}
    \caption{Robustness of semantic space STNs to the number of sampled trajectories. The global structure remains stable as the run count varies. Global efficiency is reported per panel.}
    \label{fig:runs_sweep}
\end{figure}

Another notable property of semantic space STNs is their robustness to the number of runs used in their construction. Figure~\ref{fig:runs_sweep} provides two examples showing that the overall structure of the network remains stable when the number of sampled trajectories varies. Global efficiency is reported in the figure to provide quantitative support for this observation. This stability suggests that the induced topology captures persistent aspects of the underlying optimization dynamics and indicates that semantic space STNs are structurally reliable.

\section{Analyzing Neural Network Generalization}
\label{sec:generalization}

Given a training dataset $\mathcal{D}={(x^{(i)}, y^{(i)})}_{i=0}^{N}$, supervised learning aims to learn a hypothesis $f:\mathbb{R}^d \rightarrow \mathcal{Y}$ that can predict the label $y$ associated with previously unseen covariates $x$, where $\mathcal{Y} = \{1, \dots, K\}$ for classification and $\mathcal{Y} = \mathbb{R}$ for regression. The ability of the learned hypothesis to make accurate predictions on previously unseen data drawn from the same underlying distribution as the training data is referred to as \emph{generalization}. $f$ is chosen from the hypothesis space $\mathcal{H}$ using Empirical Risk Minimization (ERM)~\cite{vapnik92}. 
\emph{Empirical risk} 
measures the average loss over the training dataset and serves as a proxy for the \emph{expected risk}, which measures the average loss under the true (and generally unknown) data distribution and reflects the model’s performance on future samples.


Traditional machine learning theory suggests that controlling 
the hypothesis space $\mathcal{H}$ makes it possible to strike a balance between underfitting and overfitting. Functions within $\mathcal{H}$ must be sufficiently expressive to capture the underlying function governing the data, yet sufficiently constrained to avoid fitting spurious patterns or noise in the training set that could impair generalization (\textit{bias–variance trade-off})~\cite{belkin2019reconciling}. 
Statistical learning theory
perspective~\cite{vapnik95} 
provides uniform convergence bounds of the schematic form:
{\small
\begin{equation*}
    \underbrace{R(f)}_{\text{Expected risk}}
    \;\le\;
    \underbrace{\hat{R}(f)}_{\text{Empirical risk}}
    \;+\;
    \mathcal{O}\!\left(\sqrt{\frac{\mathfrak{C}(\mathcal{H})}{N}}\right)
\end{equation*}
}

Where $N$ is the number of samples in the training set and  $\mathfrak{C}(\mathcal{H})$ is the capacity of $\mathcal{H}$ quantified via 
statistical complexity measures.

Interestingly, despite their high capacity and near-perfect fit to the training data, neural networks often generalize well. This appears paradoxical under the lens of Rademacher complexity: a hypothesis class $\mathcal{H}$ expressive enough to memorize labels drawn independently of the inputs attains a complexity approaching its maximum, $\mathfrak{R}_{\mathcal{D}}(\mathcal{H}) \approx 1$, which renders the bound above vacuous (see Section 2.2 of~\cite{zhang2017rethinking}).
Zhang et al.~\cite{zhang2017rethinking,zhang2021stillrethinking} explore this apparent paradox with a concrete example: standard architectures trained with stochastic gradient descent (SGD) can drive the training error to zero on randomly labeled data, memorizing the dataset with only a modest increase in training time, and doing so whether or not explicit regularization (weight decay, dropout, data augmentation) is applied. In particular, for a fixed architecture, the set of interpolating solutions is identical across the true-label and random-label regimes, so any uniform complexity measure assigns them the same capacity, even if their test behavior diverges completely. What distinguishes the two cannot, then, be the hypothesis class in isolation; it must be the interaction between 
architecture, 
data distribution, and 
optimization algorithm.

These observations reframed generalization as a property of the learned solution and the optimization trajectory, rather than of $\mathcal{H}$ alone, and motivated a large body of work on the implicit bias of gradient-based optimization. The interpretation of these phenomena, however, remains actively debated. For instance, \cite{wilson2025notsomysterious} argues that benign overfitting, double descent, and the success of overparameterization are neither unique to neural networks nor inexplicable, and can be characterized through PAC-Bayes and countable-hypothesis bounds. 
Whether or not one regards neural networks' generalization as ``mysterious'', most research on the topic studies generalization through static and local results: \cite{keskar2017largebatch} links generalization to the flatness of the reached minimum (large-batch SGD converging to sharper, worse-generalizing minima), while \cite{draxler2019essentially} and \cite{garipov2018loss} show that distinct solutions are connected by paths of near-constant low loss. 

We propose that generalization is ultimately a behavioral phenomenon: two networks may occupy very different regions of the parameter space while making nearly identical predictions, due to natural symmetries and redundancies in the search space of network parameters. We hypothesize that semantic search trajectories capture aspects of the generalization that are not apparent in the parameter space. STNs provide a graph-based representation of the regions explored during learning and of the pathways connecting them, making it possible to analyze the global organization of the semantic landscape induced by training and complementing approaches that focus on the local properties of the minima found by optimization.

\textbf{Experimental setup}
All experiments train MLPs (see Table~\ref{tab:used_models}) on the fold 0 train/test split of the benchmark introduced in Table~\ref{tab:datasets}. To sample distinct minima, we repeat each run $R=20$ times, varying the random seed of Kaiming weight initialization while holding the data, labels, and all optimization hyperparameters fixed, as in Section~\ref{sec:algo_comparaison}. Each run, therefore, solves the same learning problem from a different starting point in the parameter space.

\begin{figure}[htb]
    \centering
    \begin{subfigure}[t]{0.49\textwidth}
        \centering
        \includegraphics[width=\linewidth]{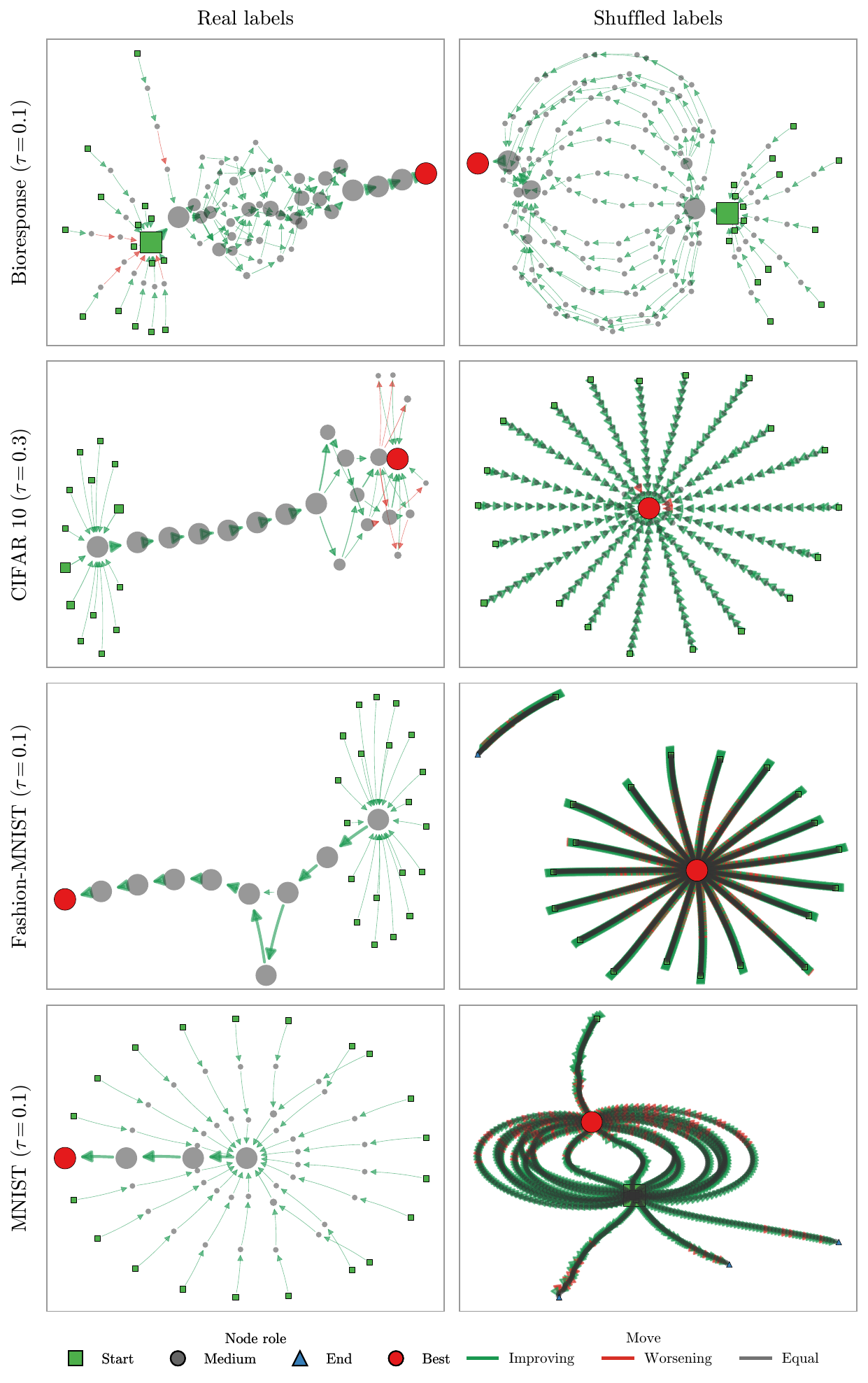}
        \caption{Classification.}
        \label{fig:real_vs_shuffled_classification}
    \end{subfigure}
    \hfill
    \begin{subfigure}[t]{0.49\textwidth}
        \centering
        \includegraphics[width=\linewidth]{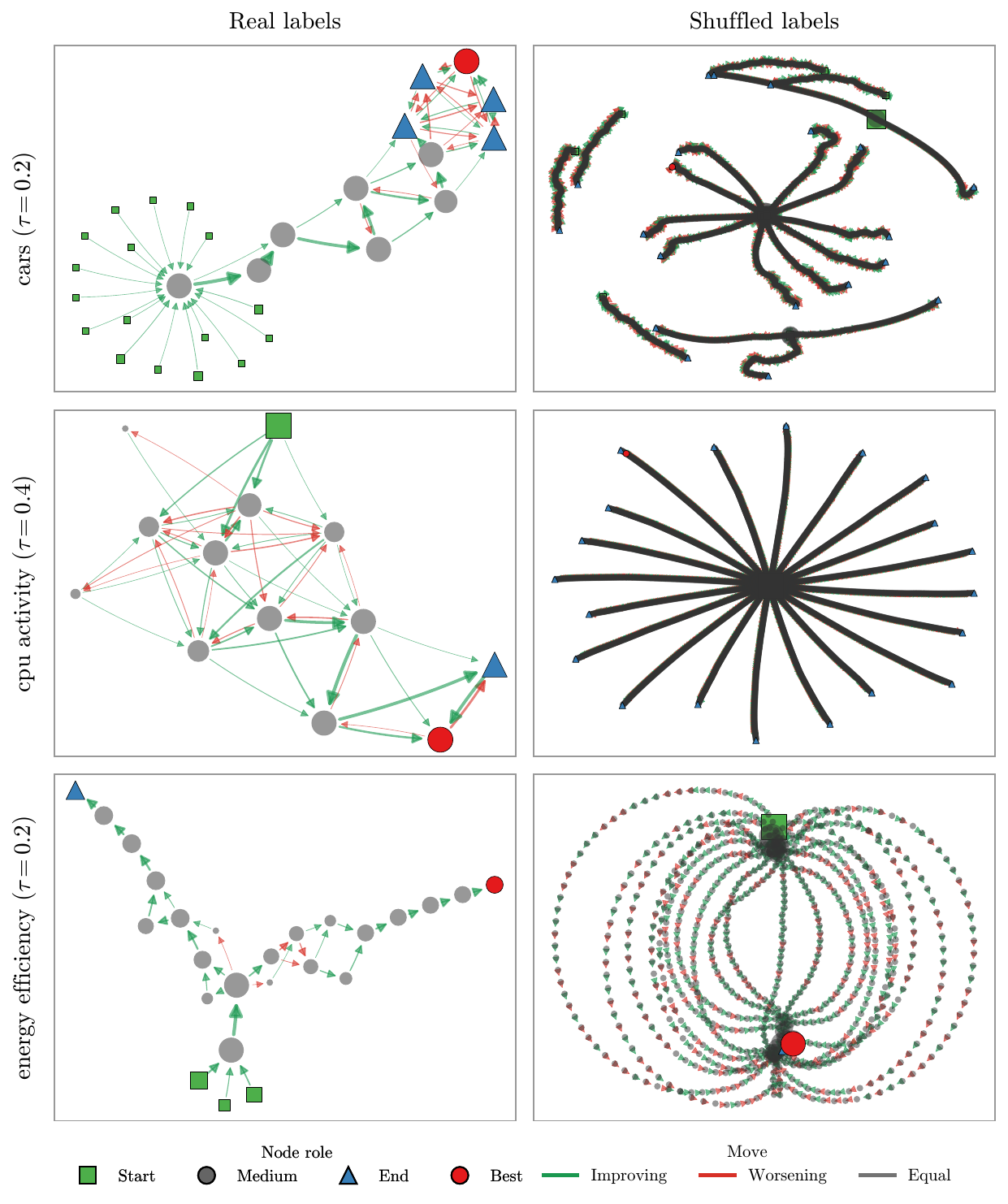}
        \caption{Regression.}
        \label{fig:real_vs_shuffled_regression}
    \end{subfigure}
    \caption{Semantic space STNs for MLPs trained under true and shuffled labels, over 20 runs. Real labels produce a funnel topology converging to a single best node or a tightly connected terminal cluster. Shuffled labels yield isolated trajectories.}
    \label{fig:combined}
\end{figure}

Following the randomization tests of Zhang et al.~\cite{zhang2017rethinking}, we propose two experiments to investigate whether a semantic space STN captures differences between the memorization and generalization regimes. First, we compare training with the true labels against training with completely shuffled labels. Second, for a single dataset, we replace a fraction of the labels with labels sampled from a random distribution.

For the first experiment, we contrast two conditions with identical inputs that differ only in the training targets: $\mathrm{real}$ (the original targets), and $\mathrm{shuffled}$ (a random permutation of the labels that destroys all input–label dependencies). The permutation is drawn once per dataset and stored, so it is shared across all runs of a given condition. Naturally, for this second case, the generalization error converges to that of random guessing.

For the second experiment, we focus on the MNIST (784) dataset and control the amount of information in the training labels through a corruption level $p\in\{0,0.2,0.4,0.6,0.8,1\}$. A fraction $p$ of the training labels is replaced by labels drawn from a uniform distribution over the classes, interpolating between the real task ($p = 0$) and fully random labels ($p = 1$). The corrupted labeling is sampled once and stored, so all $R$ initializations memorize the \emph{same} labels and differ only in their optimization trajectories.

\textbf{STNs Analysis}
%
%
Figures \ref{fig:real_vs_shuffled_classification} and \ref{fig:real_vs_shuffled_regression} show pronounced differences between the STNs produced under true and shuffled labels. When training on datasets with meaningful covariate–label relationships, the optimization process exhibits the funnel behavior described in Section \ref{sec:algo_comparaison}: trajectories in semantic space rapidly merge into a small set of interconnected paths and ultimately converge to either a single best node \stnbest or a tightly connected group of terminal nodes (\stnend and \stnbest). In contrast, with shuffled labels, the trajectories remain mutually disjoint and often converge to isolated end nodes \stnend, resulting in a star-shaped topology.


To further investigate the structural differences between STNs obtained from standard training and label shuffling, we computed a broad collection of classical graph-theoretic descriptors. We, then, examined which measures most consistently differentiate the resulting networks across datasets. Table~\ref{tab:graph_stats} reports a subset of metrics exhibiting the strongest separation, including graph density, global efficiency, maximum closeness centrality, maximum PageRank, maximum in-degree centrality, and standard deviation of node degree.

\begin{table}[h]
    \centering
    \caption{Graph structural measures of semantic space STNs trained with real and shuffled labels}
    \label{tab:graph_stats}
    \setlength{\tabcolsep}{4pt}
    \resizebox{0.85\textwidth}{!}
    {
    \begin{tabular}{l l c c c c c c}
        \toprule
        Dataset & Label & Density & Global Eff. & Closeness$_{\max}$ & PageRank$_{\max}$ & InDeg. Centr.$_{\max}$ & Std. Degree \\
        \midrule
        \multirow{2}{*}{Bioresponse}
            & random & $0.0049$ & $0.1759$ & $0.1234$ & $0.0560$ & $0.0901$ & $1.9166$ \\
            & real   & $0.0157$ & $0.2880$ & $0.2699$ & $0.0665$ & $0.2245$ & $3.0227$ \\
        \midrule
        \multirow{2}{*}{CIFAR 10}
            & random & $0.0020$ & $0.0500$ & $0.0757$ & $0.0424$ & $0.0386$ & $1.1640$ \\
            & real   & $0.0338$ & $0.3095$ & $0.4565$ & $0.1304$ & $0.4565$ & $3.9315$ \\
        \midrule
        \multirow{2}{*}{Fashion-MNIST}
            & random & $0.0003$ & $0.0091$ & $0.0115$ & $0.0052$ & $0.0059$ & $0.5889$ \\
            & real   & $0.0250$ & $0.3660$ & $0.5854$ & $0.1115$ & $0.5854$ & $3.7477$ \\
        \midrule
        \multirow{2}{*}{MNIST (784)}
            & random & $0.0005$ & $0.0274$ & $0.0166$ & $0.0078$ & $0.0099$ & $1.2031$ \\
            & real   & $0.0112$ & $0.2908$ & $0.4775$ & $0.1950$ & $0.2955$ & $2.7231$ \\
        \midrule
        \multirow{2}{*}{cars}
            & random & $0.0002$ & $0.0053$ & $0.0018$ & $0.0002$ & $0.0007$ & $0.1845$ \\
            & real   & $0.0647$ & $0.4193$ & $0.6176$ & $0.0967$ & $0.6176$ & $5.4521$ \\
        \midrule
        \multirow{2}{*}{cpu activity}
            & random & $0.0001$ & $0.0071$ & $0.0003$ & $0.0001$ & $0.0001$ & $0.1969$ \\
            & real   & $0.4095$ & $0.7333$ & $0.7000$ & $0.1447$ & $0.7143$ & $6.2382$ \\
        \midrule
        \multirow{2}{*}{energy efficiency}
            & random & $0.0010$ & $0.0493$ & $0.0466$ & $0.0312$ & $0.0063$ & $0.7698$ \\
            & real   & $0.0595$ & $0.3822$ & $0.2074$ & $0.0785$ & $0.1481$ & $2.0064$ \\
        \bottomrule
    \end{tabular}
    }
\end{table}

Although we do not claim that these metrics are optimal descriptors of semantic space STNs, they exhibit a consistent capability to discriminate between training regimes. Across all datasets, STNs obtained with real labels tend to be denser, more efficient, and more centralized (see Table~\ref{tab:graph_stats}), whereas those obtained with shuffled labels consistently exhibit lower centrality and connectivity measures, suggesting a less structured exploration of the semantic space.

Following both qualitative and quantitative analyses of STNs under the two training regimes, we formulate the hypothesis that the different structures arise naturally from the presence of an underlying function governing the relationship between covariates and labels. The intuition is that when such a function exists, optimization discovers intermediate states that solve subsets of the data in a structured, incremental manner. This leads to semantic vectors that become closer early in training and remain more tightly clustered toward convergence in semantic space.

In contrast, when the underlying information is destroyed through label shuffling, knowledge acquired from fitting individual examples is not transferable across the samples. As a result, optimization trajectories become more fragmented and disjoint, yielding STNs that are sparser and less connected.

\begin{figure}[htb]
    \centering
    \includegraphics[width=0.8\textwidth]{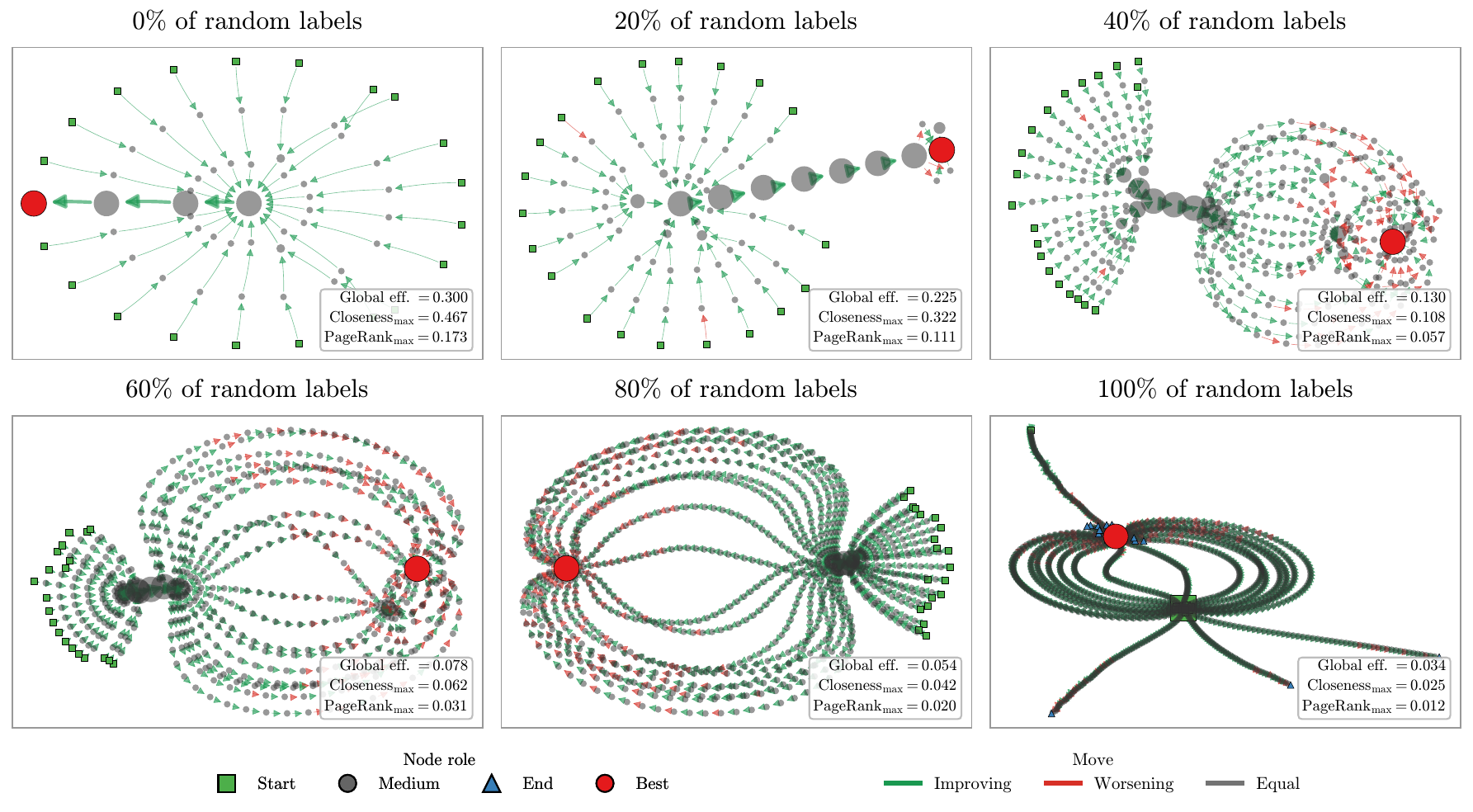}
    \caption{Progressive label corruption on MNIST (784), over 20 runs. As the corruption level $p$ increases, the STN transitions from the dense, centralised structure characteristic of the generalizing regime to increasingly isolated trajectories.}
    \label{fig:corruption_sweep}
\end{figure}

Figure~\ref{fig:corruption_sweep} shows the semantic space STNs obtained from step-by-step degradation. The evolution of STNs shows how the decrease in the amount of available information progressively transforms the graph, turning the patterns we identified as characteristic of the generalization regime into increasingly isolated trajectories. Similarly, the legends report global efficiency, maximum closeness, and maximum PageRank, which decrease monotonically.

\section{Conclusions and Perspectives}

We present a methodology for constructing Search Trajectory Networks in high-dimensional semantic spaces and apply it to classification and regression tasks, showing that our method recovers known qualitative differences between learning algorithms. Our work proposes an alternative approach to analyzing the learning behavior of machine learning algorithms, bridging complementary techniques and perspectives from different scientific communities. To further assess semantic space STNs, we compare standard training with the randomization regime of~\cite{zhang2017rethinking}. Across all datasets, models trained on structured labels yield consistently denser, more efficient, and more centralized networks than those trained on shuffled labels, with graph-theoretic descriptors cleanly separating the two regimes. These differences support a behavioral view of generalization and position semantic space STNs as a potential complement to established loss-landscape analyses.


Future work should investigate the relationship between graph-theoretic descriptors and established measures of generalization, extend the framework to a broader range of architectures, tasks, and scalable settings (including multi-class symbolic regression), and improve scalability by automating design choices such as clustering threshold selection.

\bibliographystyle{splncs04}
\bibliography{references}

\appendix

\section{Graph metrics}
Throughout this paper, we use several graph metrics as quantitative insights about semantic space STNs. This appendix provides a brief description of them and offers additional comments.
For an arbitrary graph $G(N, E)$, with a set of nodes $N$ and a set of edges $E$, metrics used during this paper can be defined as follows:

\subsection{Density}
The density of a graph quantifies how many edges are present compared to the maximum possible number of edges. For a simple directed graph, it is defined as:
\begin{equation*}
    D = \frac{|E|}{|N|(|N|-1)}
\end{equation*}
where $|N|$ is the number of nodes and $|E|$ is the number of edges.

\subsection{Global efficiency}
The global efficiency of a graph is defined as the average of the inverse shortest-path distances between all pairs of vertices~\cite{igraph2025}.
\begin{equation*}
    E = \frac{1}{|N|(|N|-1)} \sum_{i \neq j} \frac{1}{d(i,j)}
\end{equation*}
where $d(i,j)$ denotes the shortest path distance between nodes $i,j \in N$.

\subsection{Maximum closeness centrality}
Closeness centrality measures how close a node is to all other nodes in the graph. For a node $i$, it is defined as:
\begin{equation*}
    C(i) = \frac{|N|-1}{\sum_{j \neq i} d(i,j)}
\end{equation*}
The reported metric corresponds to the maximum closeness centrality over all nodes:
\begin{equation*}
    C_{\max} = \max_{i \in N} C(i)
\end{equation*}

\subsection{Maximum PageRank}
PageRank assigns an importance score $PR(i)$ to each node $i$ based on the structure of incoming links~\cite{brin1998pagerank}. In this work, PageRank is computed as implemented in NetworkX~\cite{networkx2008}, following the classical random-surfer model.

For a node $i \in N$, the PageRank score is defined as:
\begin{equation*}
    PR(i) = \frac{1-d}{|N|} + d \sum_{j \in \mathcal{N}^{in}(i)} \frac{PR(j)}{\deg^{out}(j)}
\end{equation*}

where:
\begin{itemize}
    \item $d \in [0,1]$ is the damping factor ($d=0.85$ in our case),
    \item $|N|$ is the total number of nodes in the graph,
    \item $\mathcal{N}^{in}(i)$ denotes the set of nodes linking to node $i$,
    \item $\deg^{out}(j)$ is the out-degree of node $j$.
\end{itemize}

The first term $\frac{1-d}{|N|}$ models a uniform teleportation probability, allowing the random surfer to jump to any node in the graph. The second term corresponds to the probability of reaching node $i$ by following incoming edges, weighted by the PageRank of the source nodes and normalized by their out-degree.

The PageRank vector is computed iteratively until convergence using power iteration, as implemented in NetworkX. The reported metric corresponds to the maximum PageRank value across all nodes:
\begin{equation*}
    PR_{\max} = \max_{i \in N} PR(i)
\end{equation*}

\subsection{Maximum in-degree centrality}
In-degree centrality measures the number of incoming edges to a node. A normalized version is given by:
\begin{equation*}
    C_{\text{in}}(i) = \frac{\deg^{in}(i)}{|N|-1}
\end{equation*}
where $\deg^{in}(i)$ is the in-degree of node $i$. The reported value is:
\begin{equation*}
    C_{\text{in},\max} = \max_{i \in N} C_{\text{in}}(i)
\end{equation*}

\subsection{Standard deviation of degree}
The standard deviation of the degree distribution captures the heterogeneity of node connectivity. Let $\deg(i)$ denote the degree of node $i$ and $\mu$ the average degree:
\begin{equation*}
    \mu = \frac{1}{|N|} \sum_{i \in N} \deg(i)
\end{equation*}
Then the standard deviation is:
\begin{equation*}
    \sigma_{\deg} = \sqrt{\frac{1}{|N|} \sum_{i \in N} \left(\deg(i) - \mu\right)^2}
\end{equation*}

\section{Randomization test results table}
Table~\ref{tab:shuffle_results} reports the performance metrics for the two training regimes used in the first experiment of Section~\ref{sec:generalization}. In both regimes, the network successfully fits the training set. Generalization performance is significantly degraded when the labels are shuffled.

\begin{table}[H]
    \centering
    \caption{Performance on real and label-shuffled datasets. We report mean and standard deviation over 20 runs. Regression uses $R^2$ and classification uses accuracy.}
    \label{tab:shuffle_results}
    \setlength{\tabcolsep}{4pt}
    \begin{tabular}{l c c c c}
        \toprule
        & \multicolumn{2}{c}{Real} & \multicolumn{2}{c}{Shuffled} \\
        \cmidrule(lr){2-3} \cmidrule(lr){4-5}
        Dataset & Train & Test & Train & Test \\
        \midrule
        Bioresponse
            & $1.000 \pm 0.000$
            & $0.793 \pm 0.009$
            & $0.999 \pm 0.000$
            & $0.478 \pm 0.013$ \\
        CIFAR 10
            & $1.000 \pm 0.000$
            & $0.490 \pm 0.005$
            & $1.000 \pm 0.000$
            & $0.101 \pm 0.003$ \\
        Fashion-MNIST
            & $1.000 \pm 0.000$
            & $0.888 \pm 0.002$
            & $0.994 \pm 0.001$
            & $0.099 \pm 0.003$ \\
        MNIST (784)
            & $1.000 \pm 0.000$
            & $0.976 \pm 0.001$
            & $0.994 \pm 0.001$
            & $0.096 \pm 0.004$ \\
        \midrule

        cars
            & $0.979 \pm 0.001$
            & $0.945 \pm 0.004$
            & $0.960 \pm 0.027$
            & $-1.410 \pm 0.243$ \\
        cpu activity
            & $0.999 \pm 0.000$
            & $0.977 \pm 0.001$
            & $0.949 \pm 0.008$
            & $-0.989 \pm 0.110$ \\
        energy efficiency
            & $1.000 \pm 0.000$
            & $0.997 \pm 0.000$
            & $1.000 \pm 0.000$
            & $-0.155 \pm 0.071$ \\
        \bottomrule
    \end{tabular}
\end{table}

\section{Alternative Visualization for STNs}

An alternative and commonly used method for visualizing STNs is to position nodes along the vertical axis according to their fitness values, while maintaining the Kamada-Kawai horizontal coordinates~\cite{ochoa2014local,ochoa2021search}. While we found the Kamada-Kawai layout algorithm to provide a more intuitive representation for our analysis, we have included fitness versions of all figures using this alternative visualization for the reader's reference.

\begin{figure}[htbp]
    \centering
    \includegraphics[width=0.45\textwidth]{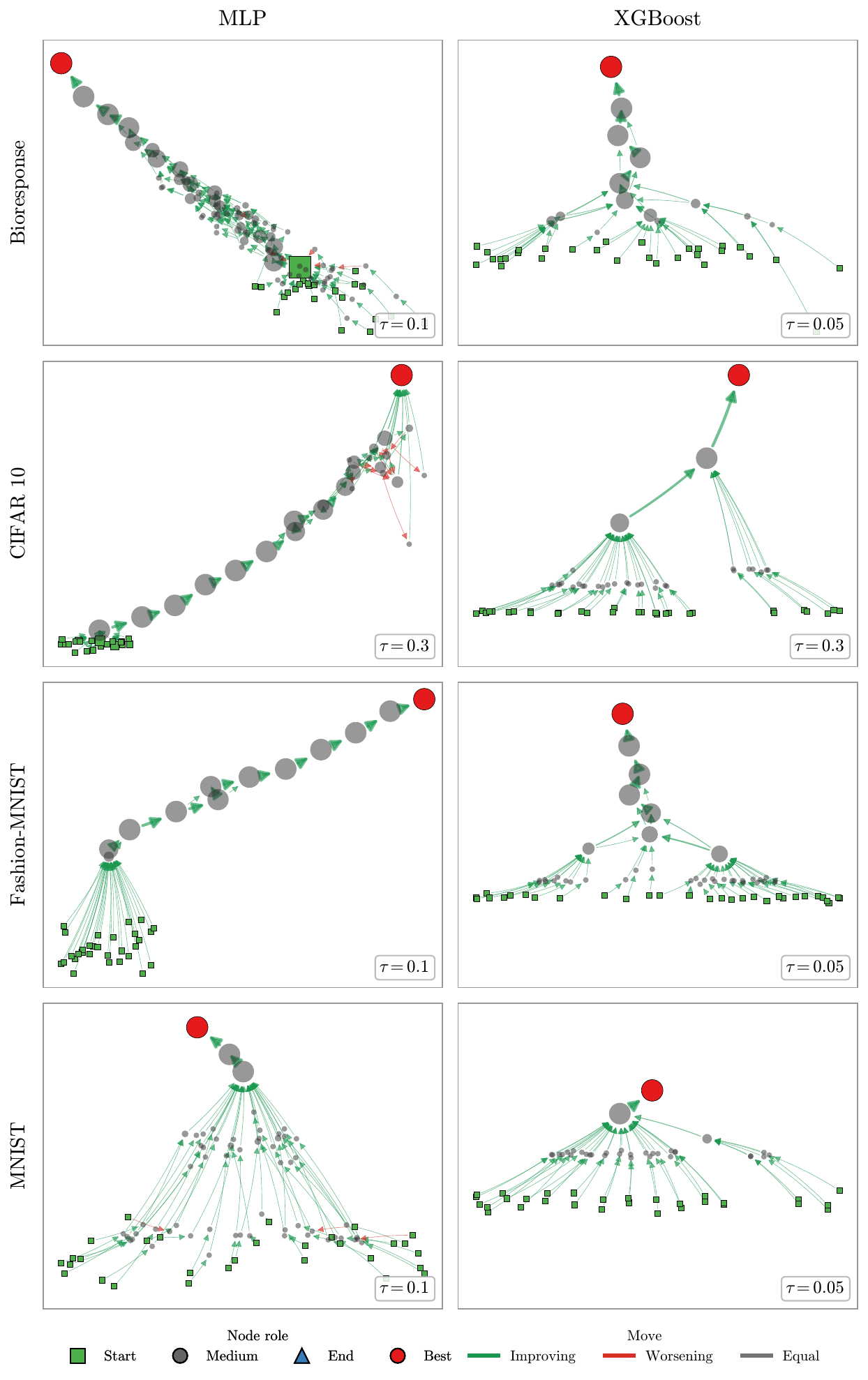}
    \caption{Semantic space STNs for MLP and XGBoost on the four classification datasets, aggregated over 30 runs. The clustering threshold $\tau$ was selected manually per dataset for readability. Fitness Layout.}
    \label{fig:comparison_classification_fitness}
    \vspace{3mm}
    \centering
    \includegraphics[width=0.63\textwidth]{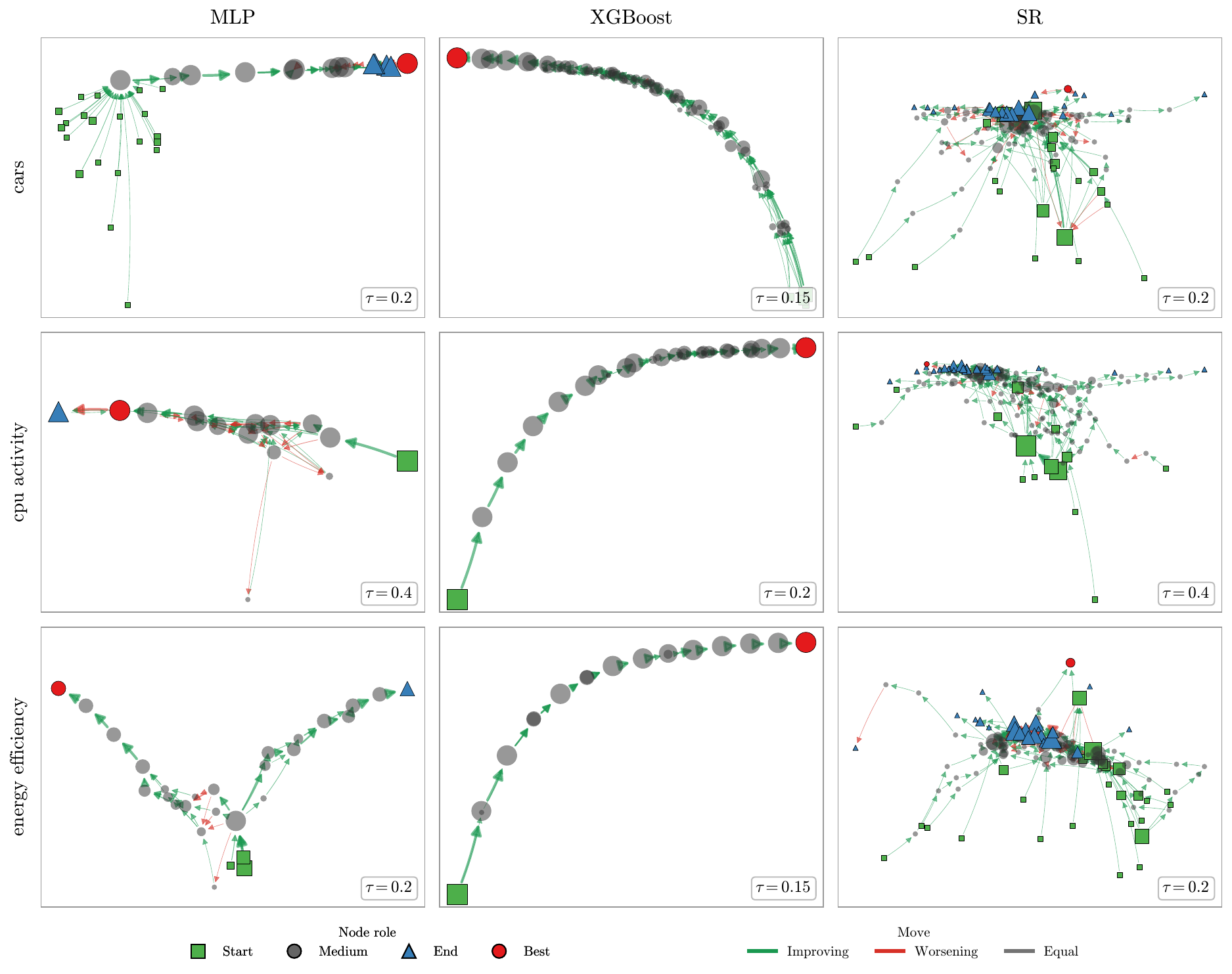}
    \caption{Semantic space STNs for MLP, XGBoost and Symbolic Regression on the three regression datasets, aggregated over 30 runs. Continuous predictions are discretised into 10 global quantile bins before Hamming-based clustering. Fitness Layout.}
    \label{fig:comparison_regression_fitness}
\end{figure}

\begin{figure}[htbp]
    \centering
    \includegraphics[width=\textwidth]{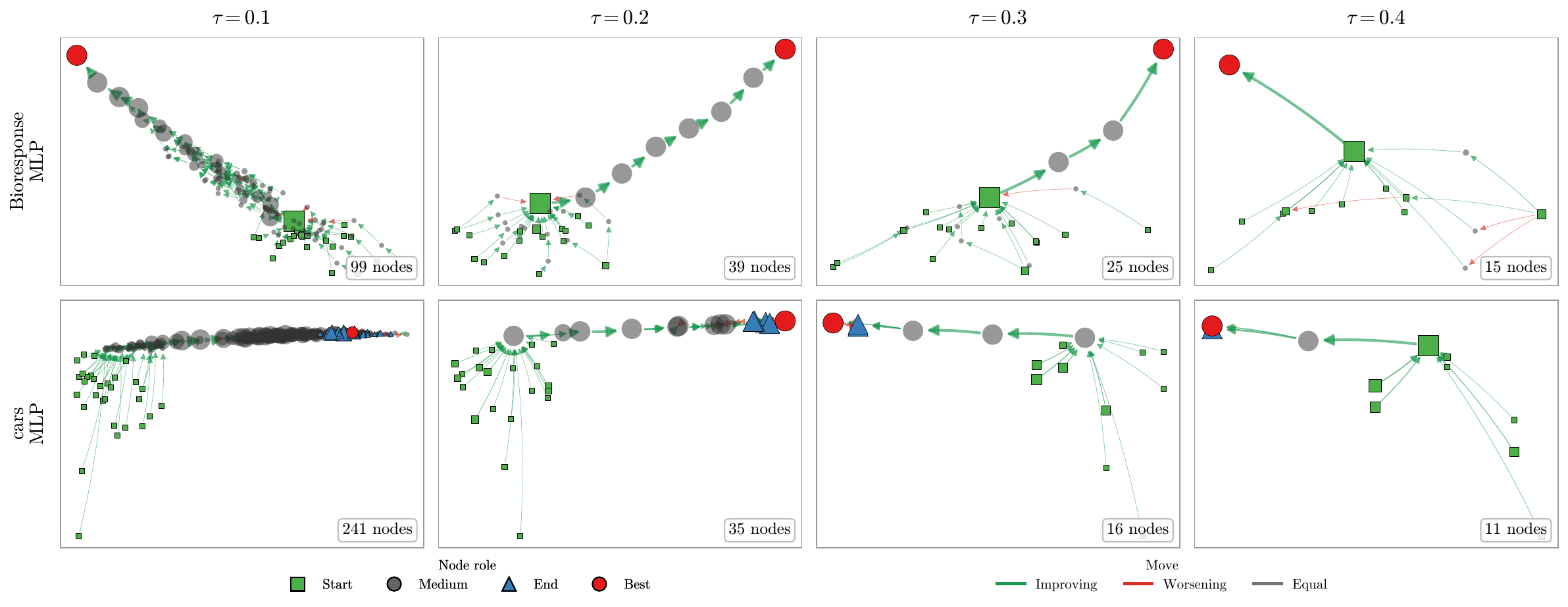}
    \caption{Effect of the agglomerative clustering threshold $\tau$ on STN granularity. Increasing $\tau$ coarsens the aggregation, merging semantic states and exposing the optimisation dynamics at progressively higher levels of abstraction. Fitness Layout.}
    \label{fig:threshold_sweep_fitness}
\end{figure}

\begin{figure}[htbp]
    \centering
    \includegraphics[width=\textwidth]{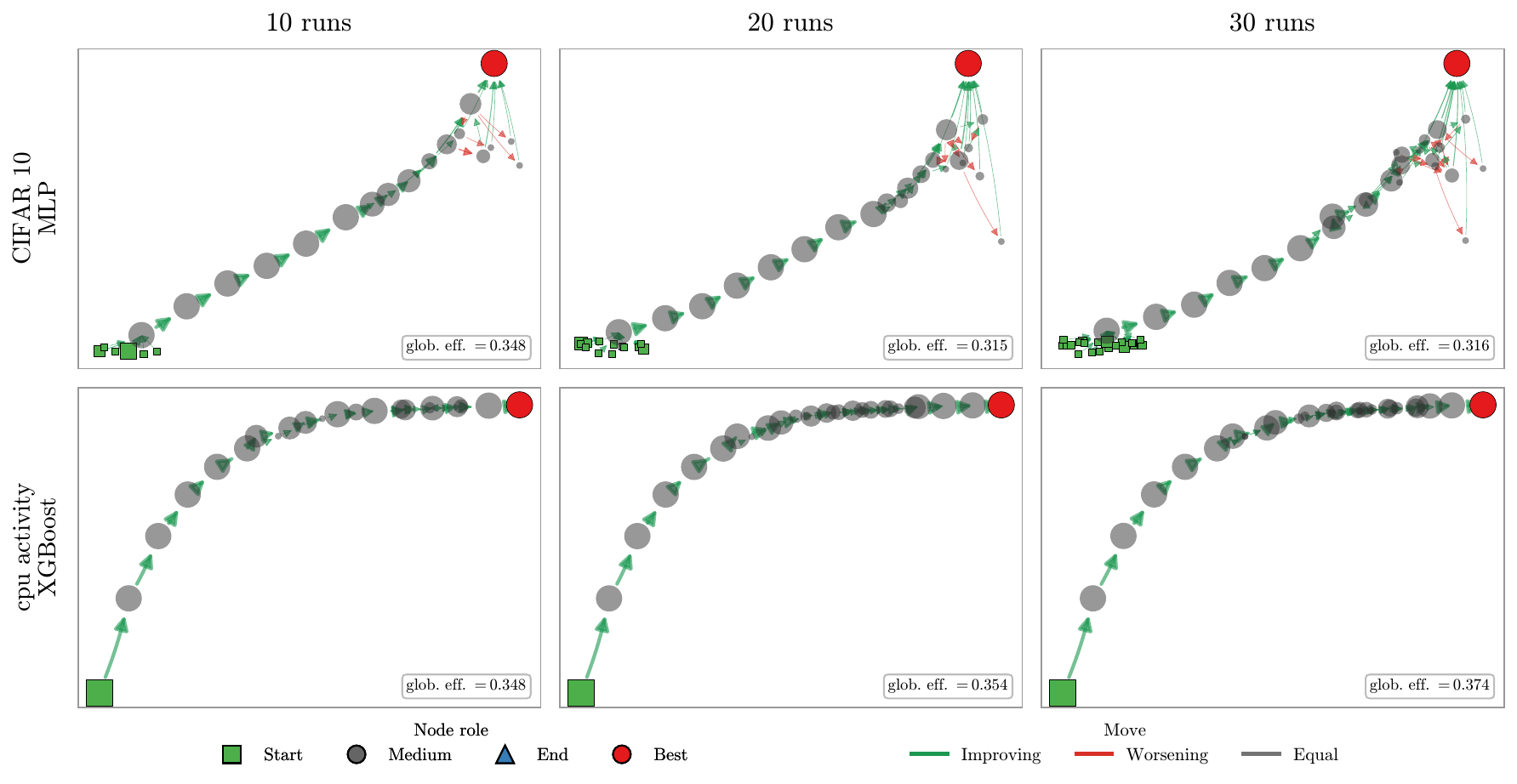}
    \caption{Robustness of semantic space STNs to the number of sampled trajectories. The global structure remains stable as the run count varies. Global efficiency is reported per panel. Fitness Layout.}
    \label{fig:runs_sweep_fitness}
\end{figure}

\begin{figure}[htbp]
    \centering
    \begin{subfigure}[t]{0.49\textwidth}
        \centering
        \includegraphics[width=\linewidth]{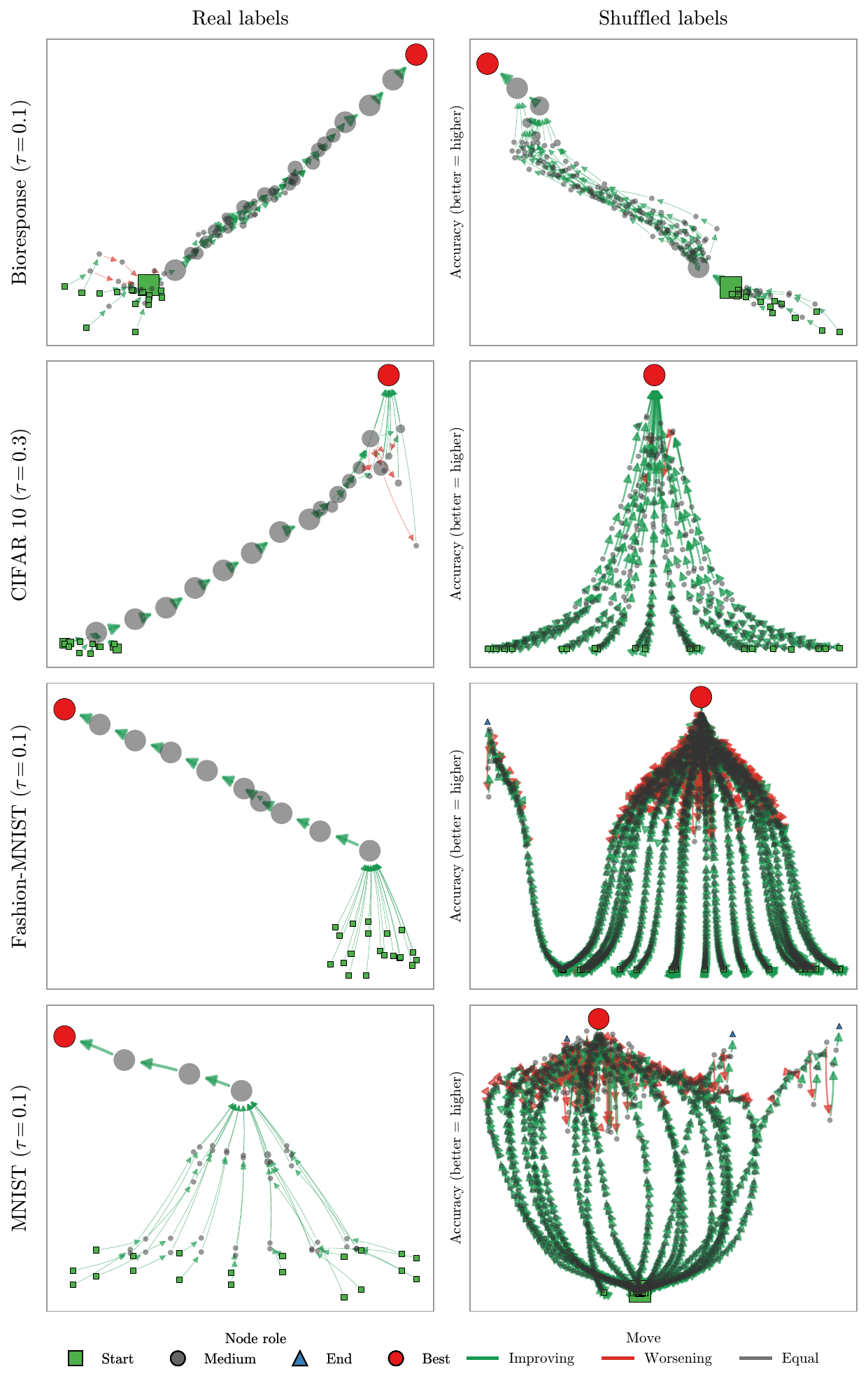}
        \caption{Classification.}
        \label{fig:real_vs_shuffled_classification_fitness}
    \end{subfigure}
    \hfill
    \begin{subfigure}[t]{0.49\textwidth}
        \centering
        \includegraphics[width=\linewidth]{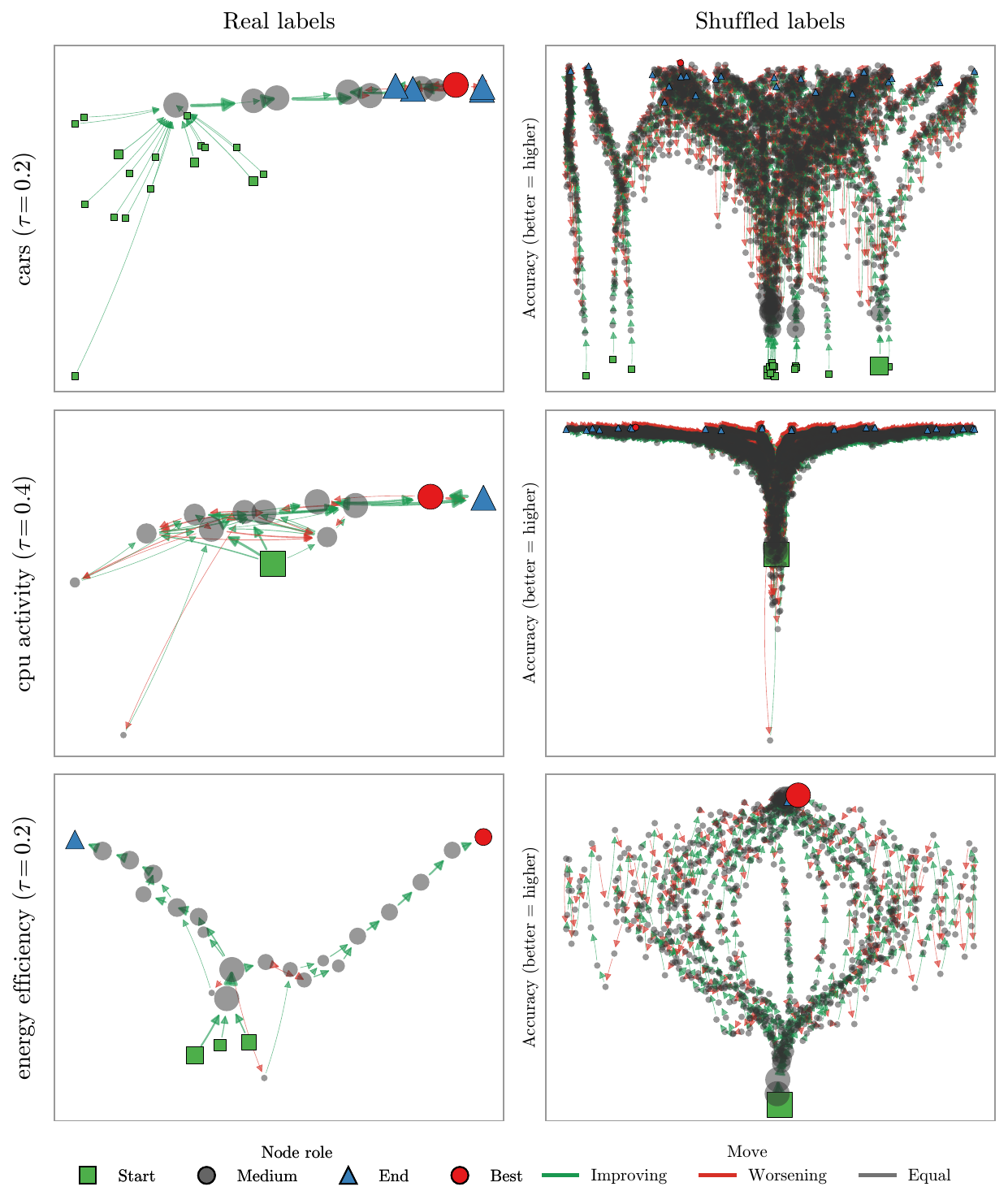}
        \caption{Regression.}
        \label{fig:real_vs_shuffled_regression_fitness}
    \end{subfigure}
    \caption{Semantic space STNs for MLPs trained under true and shuffled labels, over 20 runs. Real labels produce a funnel topology converging to a single best node or a tightly connected terminal cluster. Shuffled labels yield isolated trajectories. Fitness Layout.}
    \label{fig:combined_fitness}
\end{figure}

\begin{figure}[htbp]
    \centering
    \includegraphics[width=\textwidth]{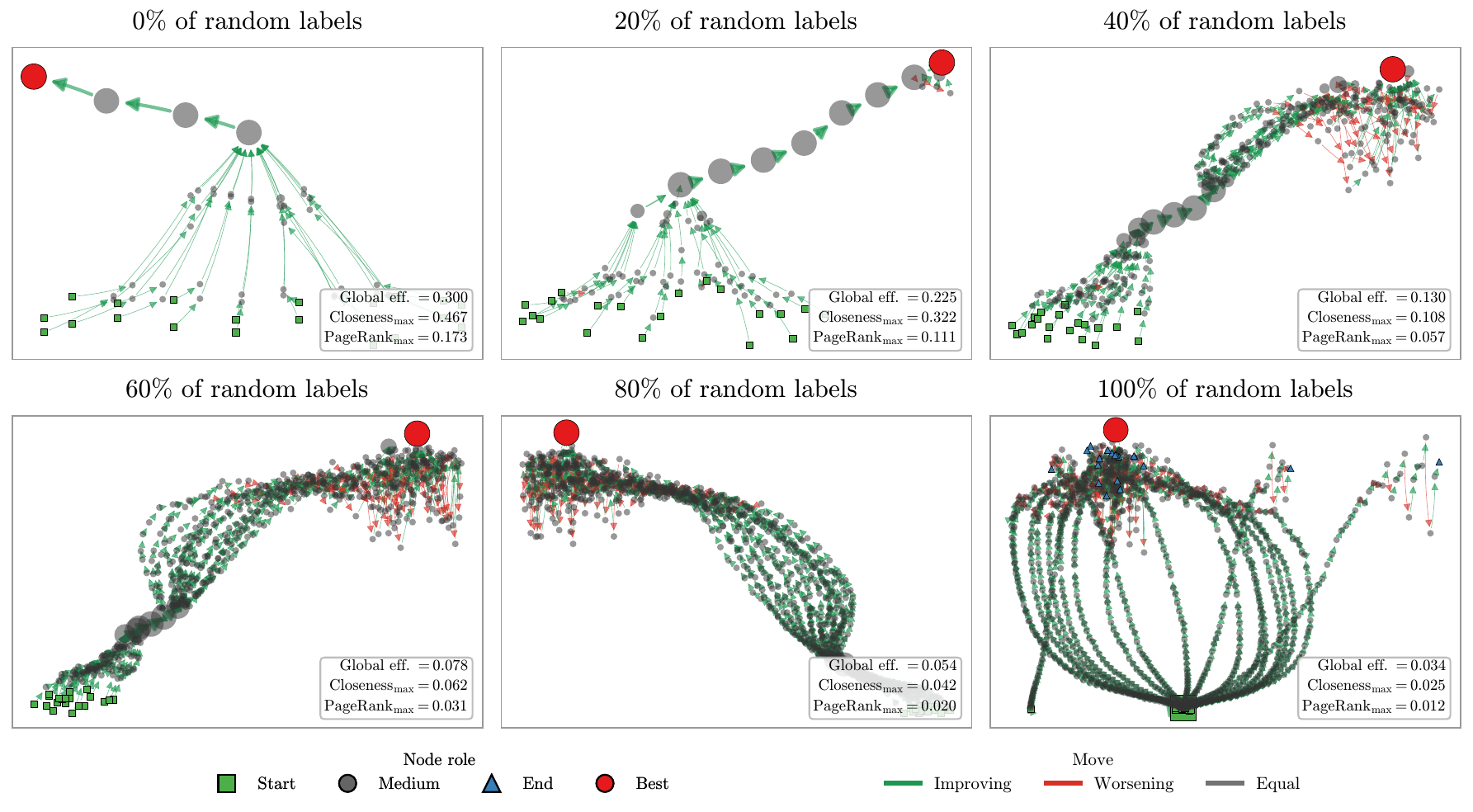}
    \caption{Progressive label corruption on MNIST (784), over 20 runs. As the corruption level $p$ increases, the STN transitions from the dense, centralised structure characteristic of the generalizing regime to increasingly isolated trajectories. Fitness Layout.}
    \label{fig:corruption_sweep_fitness}
\end{figure}

\end{document}